\documentclass[letterpaper, 10 pt, conference]{ieeeconf}  

\IEEEoverridecommandlockouts                              

\usepackage{graphics} 
\usepackage{epsfig} 
\usepackage{url}
\usepackage{subfigure}
\graphicspath{{./images/}}

\title{\LARGE \bf Robots as Actors in a Film: No War, A Robot Story}

\author{Andreagiovanni Reina, Viktor Ioannou,
  Junjin Chen, Lu Lu, Charles Kent, James A. R. Marshall
  \thanks{This work was funded by the European Research Council (ERC)
    under the European Union's Horizon 2020 research and innovation
    programme (grant agreement number 647704). J.C. received financial
    support from the University of Sheffield OnCampus Placement Scheme.}%
  \thanks{A.R, J.C., and J.A.R.M. are with the
    Department of Computer Science, University of Sheffield, S1 4DP,
    UK (email: a.reina@sheffield.ac.uk; 1099404025@qq.com;
    james.marshall@sheffield.ac.uk). V.I., L.L., and C.K.
    are with the Sheffield Institute of Arts, Sheffield Hallam
    University, S1 1WB, UK (email: victor.ioannou@gmail.com;
    lulu02261992@gmail.com; charleskentdesign@gmail.com).}%
}

\begin{document}

\maketitle
\thispagestyle{empty}
\pagestyle{empty}

\begin{abstract}
  Will the Third World War be fought by robots? This short film is a
  light-hearted comedy that aims to trigger an interesting discussion
  and reflexion on the terrifying killer-robot stories that
  increasingly fill us with dread when we read the news headlines.
  The fictional scenario takes inspiration from current scientific
  research and describes a future where robots are asked by humans to
  join the war. Robots are divided, sparking protests in robot
  society... will robots join the conflict or will they refuse to be
  employed in human warfare?
  Food for thought for engineers, roboticists and anyone imagining
  what the upcoming robot revolution could look like.
  We let robots pop on camera to tell a story, taking on the
  role of actors playing in the film, instructed through code
  on how to ``act'' for each scene.
\end{abstract}

\section{Introduction}

Among the other functions art has, art is a form of communication that
humans have used throughout history \cite{mithenBook}. Art can be
effective at provoking a reaction in its public and our art aims to
stimulate a reflection on the use of robotics in warfare. Our piece of
art comes in the form of a short film in which we touch the highly
discussed and controversial topic of military robots
\cite{Sharkey2008,Krishnan2009,Arkin2012,Galliott2016}
through a simple light-hearted comedy.

We created a film\footnote{The
  film is available online at https://youtu.be/Mt9nUDjrSqA} as we
consider it an accessible means of communication to reach the general
public. The aim is to raise public awareness on the ethical discussion
concerning the use of robots in war \cite{Sharkey2008,Arkin2012,
  Lin2014book,Veruggio2016}. While this film does not impose a view on
this matter, it pictures a fictional scenario in which robots may
reach a level of independence to allow them to make free decisions
against humans' requests for violence.

We employ the robots as the main tool in creating our film. The robots
take on the role of actors as physical bodies in front of a video
camera, interacting with each other on the stage. Robots are coded to
follow the script of each scene and react to actions of others. We
generated a screenplay-software for our robots to quickly program them
to follow the screenplay of each scene. This software, freely
available online (see Sec.~\ref{sec:software}), is not limited to this
film, but is in fact designed to allow the robots to follow generic
screenplays.


\section{Kilobot robots on stage}
\label{sec:actors}

Art and science have since always been strongly
connected~\cite{isaacson2014innovators}, therefore the rise of
robotics is also having an impact on art~\cite{herath2016robots}. In
this work, we put robots on stage. This has already been done in
various art performances, including live theatrical
shows~\cite{Demers2008,demers2016multiple}, and it traces back to the
eighteenth century with the appearance of mechanical
automata~\cite{wood2003edison,Stephens2016}. In our work, we employ
the Kilobot robots \cite{Rubenstein2012} as actors. They are small
simple robots that have been widely used in several swarm robotics
studies~\cite{Hamann2018}.

It could be argued as to whether the robots are indeed \textit{acting}
or not, as they lack of any consciousness and because there is a
decoupling between the physical body on the stage (the mechanical
robot) and the source of the actions coded by the engineer (behind the
scenes). This decoupling is still an open topic of discussion in
theatrical performances with
puppets~\cite{tillis1996actor,craig2008art}. In our work, the engineer
takes the role of the director to instruct the robots about the
screenplay. The communication happens in the form of C code uploaded
to the robots which, differently from string-operated puppets, are
autonomous agents throughout the performance. The acting performance
of a human actor is much richer and more complex than our programmed
robots, however we believe it could still represent a much simpler form
of acting. Considering robots as art performers is a recent trend that
is gaining attention \cite{herath2016robots,Demers2008} and we believe
that it is a promising direction both for engineering and art.

\begin{figure}[t]
  \centering
  \includegraphics[width=0.75\linewidth]{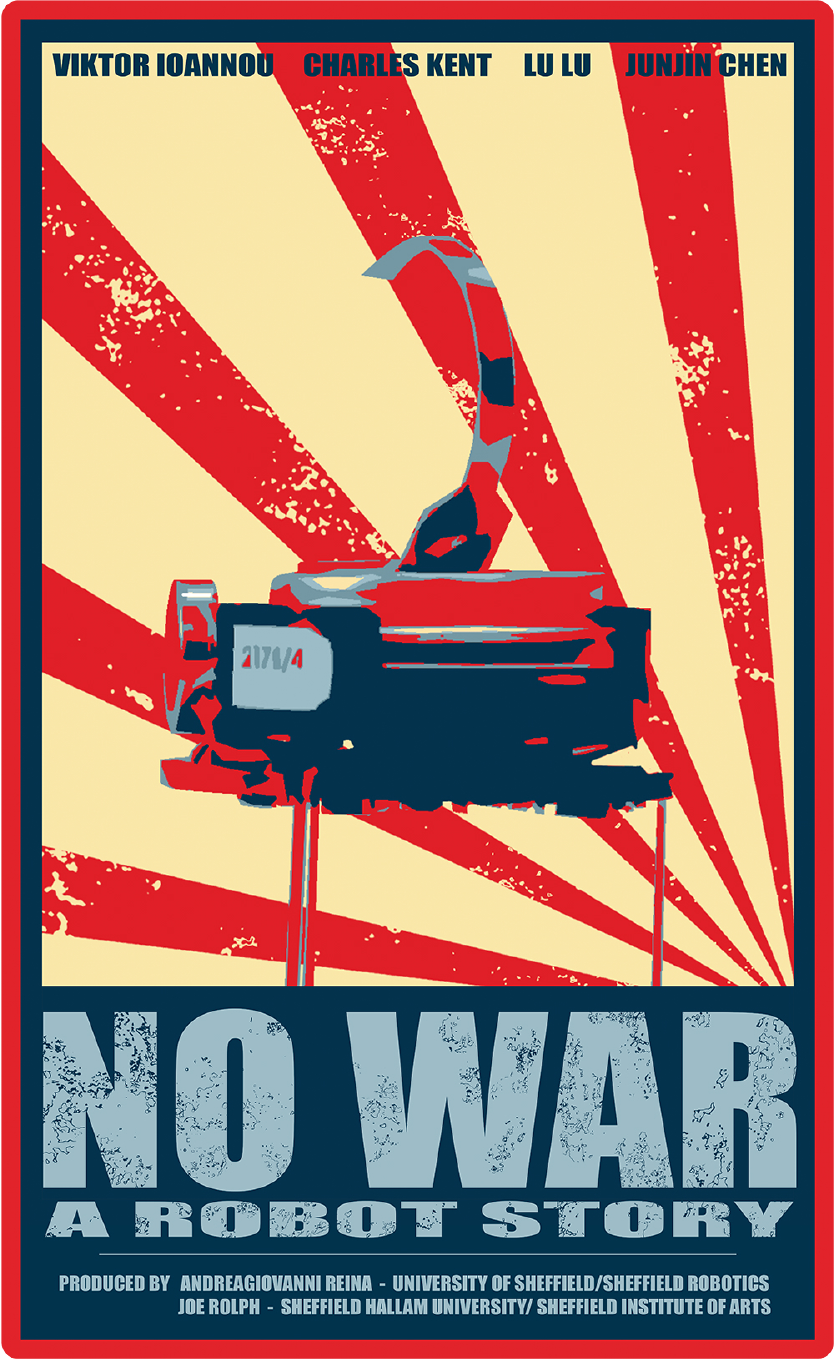}
  \caption{Poster of the film \textit{``No War: A Robot Story''}
    designed by V. Ioannou.}
  \label{fig:poster}
\end{figure}

Our art piece follows the film format by including various elements
particular to cinematography. We included a short trailer of the
film\footnote{The
  film trailer is available online at https://vimeo.com/320804400}, a
film poster (see Fig.~\ref{fig:poster}), and final sliding credits
in which robots are listed as actors and the authors of this paper
fill various film production roles (\textit{e.g.} the code programmer is
listed as the cast director, and the project supervisor as the film
producer).

Kilobots are simple robots with minimal capabilities. As they cannot
produce any synthesised speech waveform, the dialogs are therefore
implemented via speech balloons.
Still Kilobots are able to produce sounds, in fact they move powering
two vibration motors that let them vibrate on the ground causing a
buzzing sound. We exploited this peculiar vibration sound to express
the robots' intent. In particular, we used this sound to mimic laughter
and to give rhythm to the protest march (see more details in
Sec.~\ref{sec:story}).
Finally, as all Kilobots are identical robots, we added distinctive
decorations, such as a moustache or glasses, to give most of them a
recognisable character. Our goal was to stimulate robot
anthropomorphism in the audience \cite{Duffy2003,Riek2009}, as we
ultimately wish to picture the robots positively.

\section{Storyline}
\label{sec:story}

The film is composed of two phases.
In the first phase we depict a calm world with robots engaged in
various research activities. In several initial scenes we included 
``\textit{Easter-eggs}'' to acknowledge scientific work that used
Kilobots.
In the second half of the film the robots are asked to join the human
war and the calm is disrupted.

\begin{figure}[t]
  \centering
  \subfigure[Robots learn from biology]{\label{fig:library}
    \includegraphics[width=0.47\linewidth]{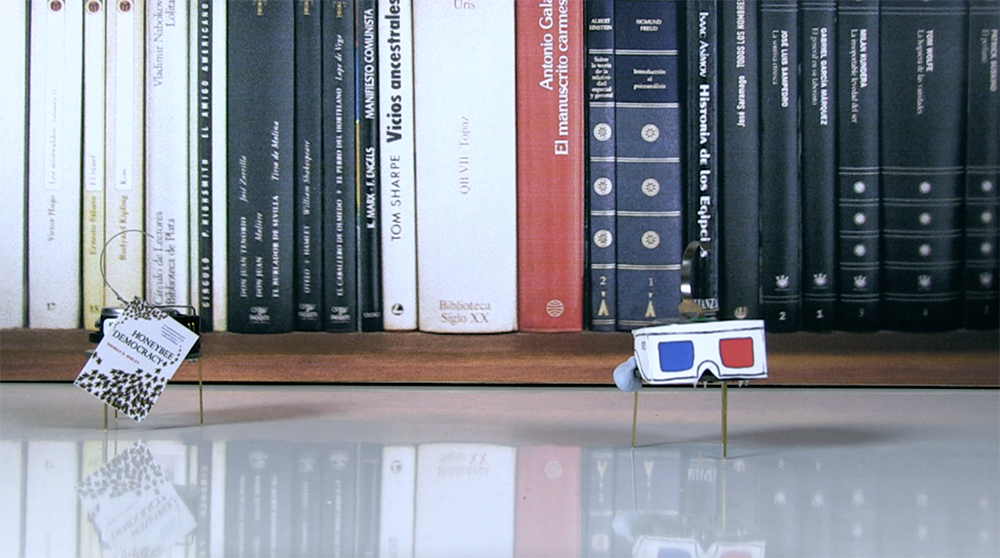}}
  \subfigure[Robots play football]{\label{fig:football}
    \includegraphics[width=0.47\linewidth]{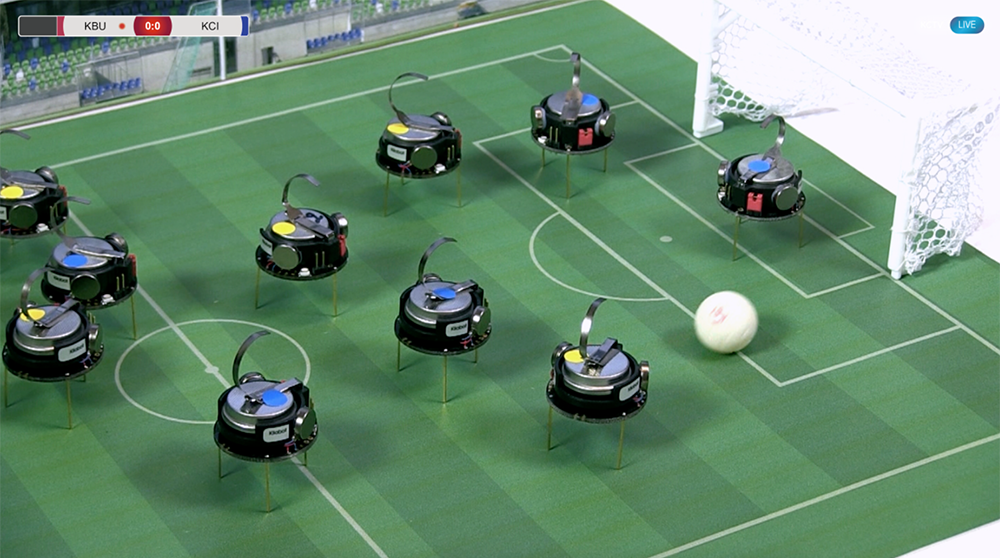}}
    \subfigure[Humans invite robots to war]{
    \label{fig:proposal}
    \includegraphics[width=0.47\linewidth]{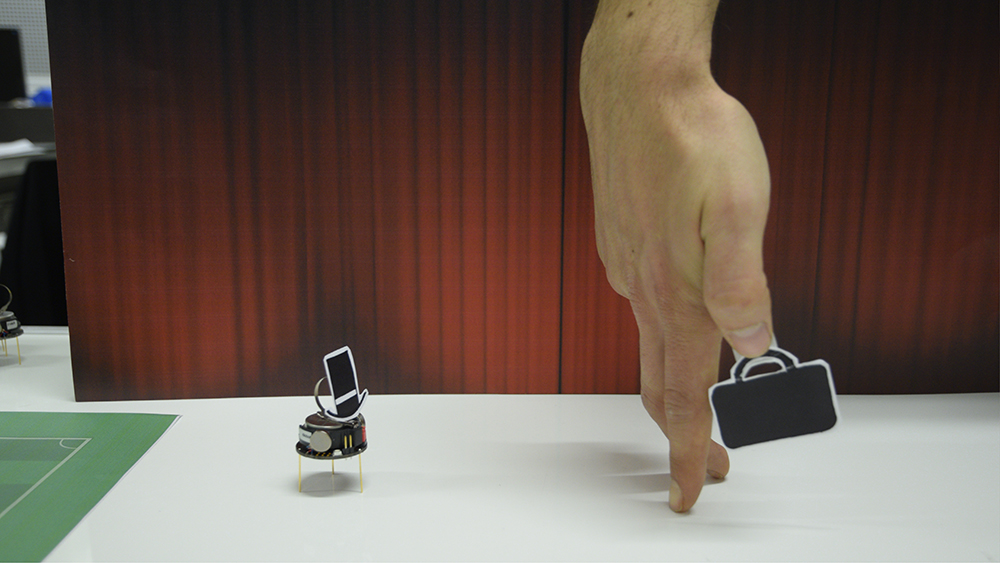}}
  \subfigure[Robots' protests spark]{\label{fig:protests}
    \includegraphics[width=0.47\linewidth]{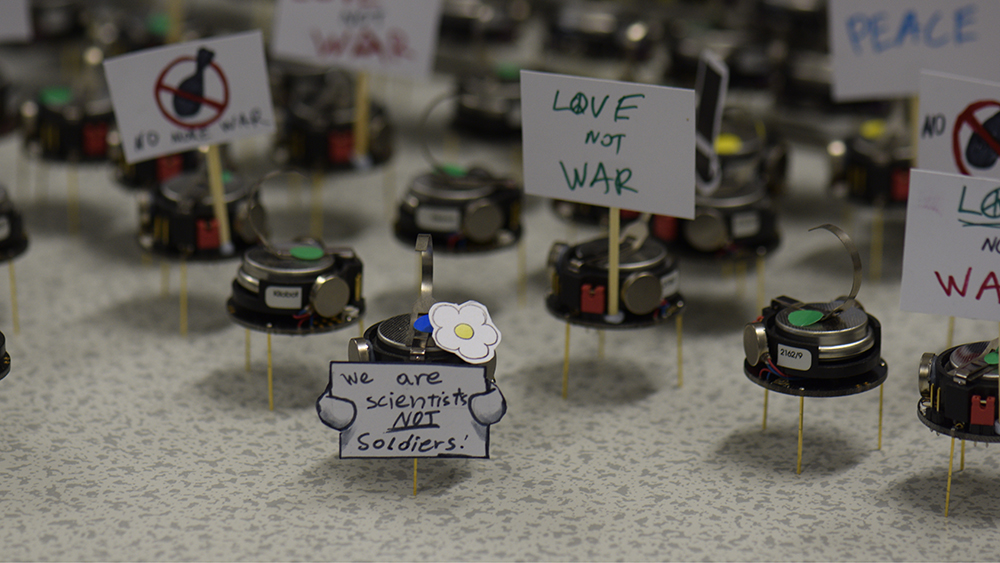}}
      \subfigure[Robots compose text with LEDs]{\label{fig:nowar}
    \includegraphics[width=0.47\linewidth]{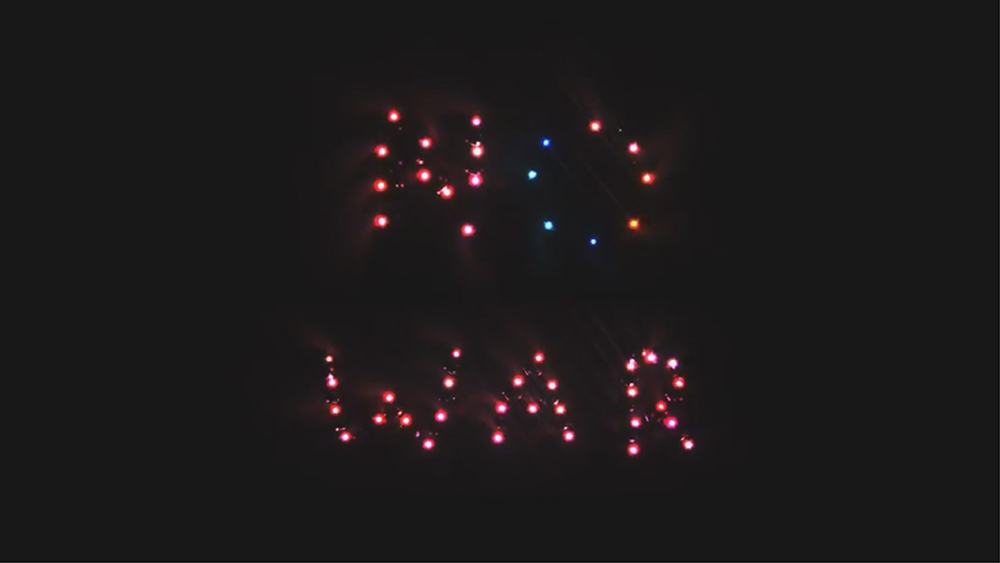}}
  \subfigure[Autonomous robots choose peace]{\label{fig:peace}
    \includegraphics[width=0.47\linewidth]{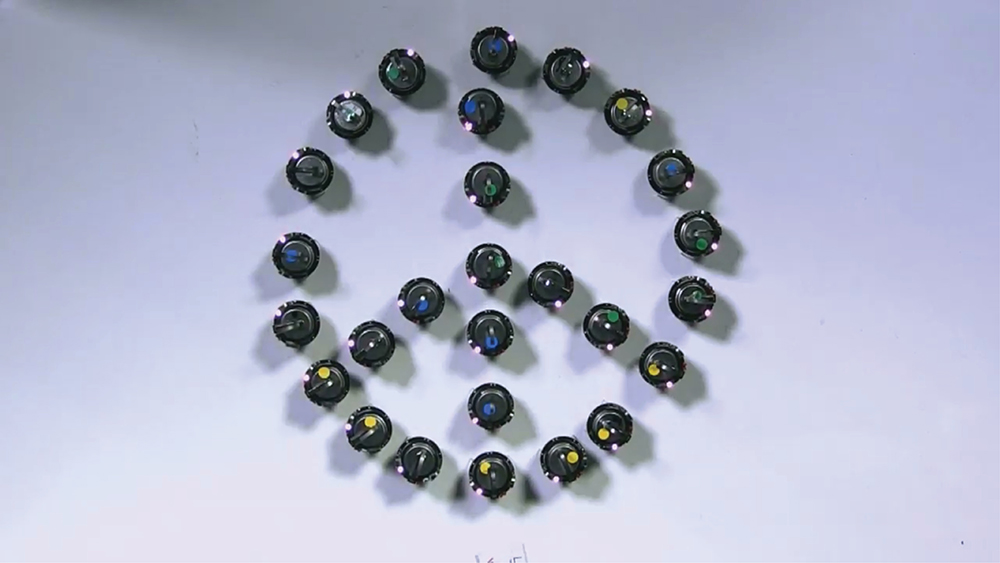}}
  \caption{Six film stills. The complete
    film is available online at https://youtu.be/Mt9nUDjrSqA
    \vspace{-0.2cm}}
\label{fig:screens}
\end{figure}

Here we describe the meaning of each scene in the first phase and the
message it aims to convey to the public.
\begin{itemize}
\item The Kilobot robot is introduced as a researcher of the
  scientific community, however, it immediately makes a grammatical
  mistake when
  it introduces itself. This simple mistake is intended to indicate the
  simplicity of these robots which are frequently subject to erratic
  or noisy behaviour.
\item ``The Kilobots are artists''. This scene self-references to this
  work in which Kilobots play as actors on the stage, and therefore
  can be classed to some extent as artists. Additionally in this
  scene, screenshots of some
  scientific studies conducted with the Kilobots are depicted as
  painted canvasses in the background, acting as hidden citations
  (Easter-eggs). The first two canvasses refer to
  the studies of~\cite{DivbandSoorati2018,Rubenstein795}; the third
  canvas is an ad-hoc image created for this work, which is generated
  through a high-exposure image of illuminated Kilobots on a random
  walk~\cite{Dimidov2016}.
\item ``Kilobots are social and like hanging out with mates'': this
  scene
  refers to the `social' nature of these robots. In fact
  they have been designed and are mostly employed in swarm robotics,
  which focuses on interactions in large group of
  individuals~\cite{Hamann2018,Brambilla2013}. The hidden citations of
  this scene are items carried by two of the Kilobots. A Kilobot
  carries a watering can to cite the
  casestudy of~\cite{antoun|etal:2016} in which robots perform the
  task of irrigating dry
  areas through a virtual watering device. The other Kilobot carries a
  basket with red mushrooms, representing a citation to the
  foraging research study
  of~\cite{FontLlenas:ANTS:2018,Talamali:SI:2019} in which
  Kilobots collect virtual items depicted as mushrooms.
\item ``Kilobots like learning and evolving'': this is a reference to the vast
  research fields in robotics, and swarm robotics, of reinforcement
  learning~\cite{panait2005cooperative,Li2011} and evolutionary
  robotics~\cite{trianni2008evolutionary,Trianni2014}. The Kilobots
  are in a library (see Fig.~\ref{fig:library}), and the hidden
  citation is one robot reading the
  book ``\textit{Honeybee Democracy}'' \cite{seeley2010honeybee} which
  inspired several behaviours that have been tested on
  Kilobots~\cite{Valentini2016,Reina:DARS:2016,Talamali:ICRA:2019}.
  The robot wearing 3D glasses refers to
  the two augmented reality systems that have been proposed in 
  recent years to allow Kilobots to see a virtual
  world~\cite{Reina:RAL:2017,Valentini2018}.
\item ``Kilobots like telling jokes in the local pub'': this scene refers
  to studies investigating the diffusion of social norms, or
  memes, into populations just through local
  interactions~\cite{Baronchelli2017}. Some of
  these studies
  have also been implemented on Kilobots to understand the role of
  a physical device implementation~\cite{Trianni:RAL:2016}.
\item ``Kilobots like going out for lunch'': this scene shows the robots
  hanging on the dock station for battery charging. The metaphor of
  describing battery charging as
  eating aims at robot anthropomorphism in order to bring the robots closer to
  the public (as discussed in
  Sec.~\ref{sec:actors}).
\item The football scene (see Fig.~\ref{fig:football}) acknowledges
  the scientific efforts that, for the past few decades, have been dedicated to the
  RoboCup~\cite{Kitano:1997}---an initiative in which research
  groups compete by enabling their robots to play football matches.
  Playing football
  requires solving several challenging tasks, such as vision, motion,
  and team coordination. Framing the research efforts onto football
  attracts public interest (and potential research funding) in
  robotics, which may otherwise be less entertaining to non-experts.
\item The dancing scene pivots on the robots' LEDs and local
  coordination between robots. While not entirely visible, the robots
  synchronise their movements through standard synchronisation
  algorithms~\cite{strogatz2004sync,strogatz1993coupled}, which have
  been also
  investigated in swarm robotics~\cite{perez2018emergence}. The robots
  also mimic dancing through an orbiting behaviour that has been
  designed in the first major work featuring Kilobots~\cite{Rubenstein795}.
\end{itemize}

The second part of the film sees humans asking Kilobots to join the
war (see Fig.~\ref{fig:proposal}). The robots show independence;
protests spark in the robot society (Fig.~\ref{fig:protests}). The
robots are against being involved in warfare. The protests are
communicated with robots moving in a rhythmic pattern which evokes
people's discontent. 


\section{Kilobot's Screenplay Software}
\label{sec:software}

We developed a software for generic screenplay. The code is open-source
and available online at
\url{https://github.com/DiODeProject/KiloArtFunctions}. The user can
specify in an intuitive text format a set of actions that a robot
should follow. The software automatically translates, in a transparent
way, the user-specified actions into C code to control the Kilobot.
The Kilobots are very simple and therefore their possible actions are
limited. In particular, a Kilobot has four actuators that can be operated:
it can move through two vibration motors, send infrared (IR) messages,
and turn on its coloured LED. The user can therefore specify actions
related to these actuators (\textit{e.g.} move left and light-up red,
or move straight and send an IR message). Each action can be repeated
for a predefined length of time, and blocks of actions repeated a
predefined number of times or until the robot senses an event. The
Kilobot only has two sensors; an ambient light sensor and an IR
receiver. For this work we only used the IR receiver, and therefore the
robot can only sense if other robots are sending IR messages
(\textit{e.g.} the user can program the Kilobot to blink red and blue
intermittently until it stops receiving neighbours' messages).

This software is quite generic and versatile for screenplays for
Kilobots as it covers the use of most onboard sensors and actuators.
Most scenes of this film have been implemented through this software,
and we believe that it can be employed by non-programmers to develop
other simple stories with the Kilobots.

\section{Take home message}
\label{sec:conclusion}

As robots are becoming more autonomous, there is large interest in the
development of ethical robots
\cite{Sharkey2008, Krishnan2009, Arkin2012, Lin2014book, Galliott2016,
  Vanderelst2018}. Discussions on robot morality and ethics are
complex and often lead to philosophical dilemmas which even humans
cannot resolve~\cite{Veruggio2016,Anderson2011}. A controversial topic
concerns how to develop ethical lethal military robots
\cite{Arkin2012,Arkin2009}, or how to completely prevent the
development of any robotic weapons~\cite{Gubrud2014,Sparrow2016} as
robot ethics may be compromised~\cite{Vanderelst2016}. We aim to raise
public awareness on this topic through a short film that pictures a
fictional distant future in which autonomous robots reach a level of
consciousness to be able to refuse any military involvement in favour
of peace. This scenario is just science fiction, and very distant from the
robotics state of art, however, as noted by others \cite{Lin2011},
Isaac Asimov predicted \textit{``It is change, continuing change,
  inevitable change, that is the dominant factor in society today. No
  sensible decision can be made any longer without taking into account
  not only the world as it is, but the world as it will be\ldots This, in
  turn, means that our statesmen, our businessmen, our everyman must
  take on a science fictional way of
  thinking''}~\cite{asimov1982science}. Our science fiction short film
allowed the robots to be the protagonists, pop on camera, and tell a
story in which robots are so intelligent that they say `No to war'; an
optimistic future.

\section*{ACKNOWLEDGMENT}
The authors would like to thank Danielle Buck for the fruitful
discussions, and acknowledge Melanie Levick-Parkin and Joe Rolph for
the administrative coordination of the SHU students.

\addtolength{\textheight}{-1cm}

\bibliographystyle{IEEEtran}  
\bibliography{ref_icrax}

\end{document}